\documentclass{article}

\usepackage{arxiv}
\usepackage{color,soul}
\usepackage{longtable}
\usepackage{adjustbox}
\usepackage[utf8]{inputenc} 
\usepackage[T1]{fontenc}    
\usepackage{hyperref}       
\usepackage{url}            
\usepackage{booktabs}       
\usepackage{amsmath,amssymb,amsthm, amsfonts} 
\usepackage{array,multirow}
\usepackage{nicefrac}       
\usepackage{microtype}      
\usepackage{lipsum}		
\usepackage{graphicx}
\usepackage{natbib}
\usepackage{doi}
\usepackage{tikz}
\usetikzlibrary{shapes, matrix, positioning}

\newcommand\blfootnote[1]{%
  \begingroup
  \renewcommand\thefootnote{}\footnote{#1}%
  \addtocounter{footnote}{-1}%
  \endgroup
}

\title{Linear Pretraining in Recurrent Mixture Density Networks}

\author{ Hubert Normandin-Taillon\\
	Department of Computer Science and Software Engineering\\
	Concordia University\\
	Montréal, Canada \\
	\texttt{hubert.normandin-taillon@mail.concordia.ca} \\
	\And
	\href{https://orcid.org/0000-0001-5097-5269}{\includegraphics[scale=0.06]{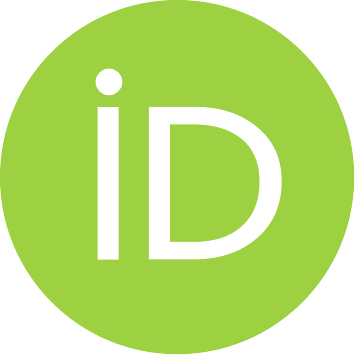}\hspace{1mm}Frédéric Godin} \\
	Department of Mathematics and Statistic\\
	Concordia Universty\\
	Montréal, Canada \\
	\texttt{frederic.godin@concordia.ca} \\
	\AND
    \href{https://orcid.org/0000-0001-6994-0146}{\includegraphics[scale=0.06]{orcid.pdf}\hspace{1mm}
	Chun Wang} \\
	Concordia Institute for Information Systems Engineering (CIISE)\\
    Concordia University\\
	Montréal, Canada \\
	\texttt{chun.wang@concordia.ca} \\
}

\renewcommand{\shorttitle}{Linear Pretraining in Recurrent Mixture Density Networks}

\hypersetup{
pdftitle={Linear Pretraining in Recurrent Mixture Density Networks},
pdfsubject={cs.LG},
pdfauthor={Hubert Normandin-Taillon, Frédéric Godin, Chun Wang},
pdfkeywords={mixture density networks, pretraining, neural networks},
}

\begin{document}
\maketitle

\begin{abstract}
We present a method for pretraining a recurrent mixture density network (RMDN). We also propose a slight modification to the architecture of the RMDN-GARCH proposed by \cite{Nikolaev2012}. The pretraining method helps the RMDN avoid bad local minima during training and improves its robustness to the \textit{persistent NaN problem}, as defined by \cite{Guillaumes2017MixtureDN}, which is often encountered with mixture density networks. Such problem consists in frequently obtaining ``Not a number" (NaN) values during training. The pretraining method proposed resolves these issues by training the linear nodes in the hidden layer of the RMDN before starting including non-linear node updates. Such an approach improves the performance of the RMDN and ensures it surpasses that of the GARCH model, which is the RMDN's linear counterpart.
\end{abstract}


\section{Introduction}
\label{intro}
Financial time series are known for their rich dynamics, which incorporates heteroskedasticity, skewness and fat tails, among other stylized facts. For such reason, distributional forecasting of stock returns is a highly non-trivial task. Multiple models have been developed to model heteroskedastic time series. Pioneering works in such strand of literature include the development of the Autoregressive Conditional Heteroskedasticity (ARCH) model \citep{engle1982} and the Generalized Autoregressive Conditional Heteroskedasticity (GARCH) model \citep{bollerslev1986}. These simple models are able to capture the persistence in the evolution of volatility present in financial time series, and to produce associated volatility clusters.
The use of Hidden Markov Models (HMM), also referred to as regime-switching models, also gained traction in the literature since the seminal work of \cite{hamilton1989new}. Such an approach involves modeling return distributions for which parameters are modulated through a Markov chain, which therefore produces mixtures of distributions. Other papers combined the two aforementioned approaches and developed regime-switching GARCH models, see for instance \cite{gray1996modeling} and \cite{klaassen2002improving}, which allow for time-varying parameters in the mixture distributions, and thus for more volatility persistence.

\blfootnote{Financial support from Mitacs, Quantolio and NSERC (Godin: RGPIN-2017-06837) is gratefully acknowledged.}

More recently, these econometric models were extended in the machine learning literature through the introduction of recurrent mixture density networks (RMDN). The model presented by \cite{schittenkopf_forecasting_2000} is a non-linear RMDN, which is extended by \cite{Nikolaev2012} who include linear nodes in the hidden layer and propose to use the RTRL algorithm for training the RMDN\citep{Nikolaev2012}. We refer to this last model as the RMDN-GARCH. RMDNs can capture the time-varying shape of the conditional distribution of a time series in a semi-parametric fashion, and represent non-linear dependencies of the mean and variance with respect to previous observations from the time series. However, mixture density networks are notoriously difficult to train and are sensitive to the initialization method; they can easily get stuck in bad local minima during training \citep{hepp_mixture_2022}.

We propose a small alteration to the architecture of the RMDN-GARCH proposed by \cite{Nikolaev2012}, which makes use of some of the proposed improvements to the MDN architecture\citep{Guillaumes2017MixtureDN}. We also propose a novel training method, which relies on a pretraining procedure, where only a subset of weights from the RMDN are updated in initial backpropagation iterations, with all other weights being temporarily frozen. Such procedures allow approaching the predictive behavior of the GARCH model in early training stages and thus to surpass it in later training iterations. Numerical experiments presented hereby show that in absence of pre-training, the RMDN-GARCH sometimes even fails to reach GARCH performance in-sample, which is indicative of bad training due to the fact that the GARCH model is nested within the RMDN-GARCH model. While our method does not guarantee convergence to a global minima, it significantly improves the robustness of the model to the initial values of the network's parameters. 

\section{The RMDN Model Trained by Backpropagation}
\label{model}
In this section, we propose a small modification to the architecture of the RMDN-GARCH model of \cite{Nikolaev2012}; such modification consists in changing the final activation function of the variance network of the RMDN, and reflects recently proposed improvements to the architecture of mixture density networks\citep{Guillaumes2017MixtureDN}. We refer to the new architecture as the ELU-RMDN. Following the work of \cite{Nikolaev2012}, such recurrent mixture density network is used to forecast the conditional density of stock returns through mixtures of Gaussian distributions. The training method that we propose also differs from the RTRL method used by \cite{Nikolaev2012}.

\begin{figure}[!h]
    \centering
    \scalebox{0.60}{
\begin{tikzpicture}[
roundnode/.style={circle, draw=black!60, fill=gray!5, very thick, minimum size=8mm},
squarednode/.style={rectangle, draw=black!60, fill=gray!5, very thick, minimum size=7mm},
smallsquarednode/.style={rectangle, draw=black!60, fill=gray!5, very thick, minimum size=5mm},
diamondnode/.style={shape=diamond, draw=black!60, fill=gray!5, very thick, minimum size=7mm},
node distance=0.75cm
]
\node[squarednode, label=south:1](input_bias) {};
\node[squarednode, label=south:$r_{t}$] (input) [right=5cm of input_bias]{};

\node[roundnode] (hidden_mix1) [above=of input_bias] {$g$};
\node[roundnode] (hidden_mix0) [above=of input_bias, left=of hidden_mix1] {$\Sigma$};
\node[roundnode] (hidden_mix2) [above=of input_bias, right=of hidden_mix1] {$g$};

\node[roundnode] (hidden_mean1) [above=of input] {$g$};
\node[roundnode] (hidden_mean0) [above=of input, right=of hidden_mix2, left=of hidden_mean1] {$\Sigma$};
\node[roundnode] (hidden_mean2) [above=of input, right=of hidden_mean1] {$g$};

\node[diamondnode, label=west:$\overline{\mu}_{t+1}$] (sum) [below right=0.3cm and 1cm of hidden_mean2] {};

\node[squarednode, label=south:$e_{t}^2$](var_input) [right=4cm of input]{};
\node[squarednode, label=south:$1$](var_input_bias) [right=of var_input]{};
\node[diamondnode, label=south east:$h_{1,t}$](hidden_input0) [right=of var_input_bias]{};
\node[diamondnode, label=south east:$h_{2,t}$](hidden_input1) [right=of hidden_input0]{};

\node[roundnode] (hidden_var0) [above=of var_input] {$\Sigma$};
\node[roundnode] (hidden_var1) [above=of var_input_bias] {$g$};
\node[roundnode] (hidden_var2) [above=of hidden_input0] {$\Sigma$};
\node[roundnode] (hidden_var3) [above=of hidden_input1] {$g$};

\node[roundnode] (output_mix0) [above right=1cm and 0.25cm of hidden_mix0] {$s$};
\node[roundnode] (output_mix1) [above right=1cm and 0.25cm of hidden_mix1] {$s$};
\node[smallsquarednode, label=south :$1$](hidden_mix_bias) [below left=0.7cm and 1.5cm of output_mix0]{};
\node[roundnode] (output_mean0) [above right=1cm and 0.25cm of hidden_mean0] {$.$};
\node[roundnode] (output_mean1) [above right=1cm and 0.25cm of hidden_mean1] {$.$};
\node[smallsquarednode, label=south:$1$](hidden_mean_bias) [below left=0.7cm and 1.5cm of output_mean0]{};
\node[roundnode] (output_var0) [above=0.75cm of hidden_var1] {e};
\node[roundnode] (output_var1) [above=0.75cm of hidden_var2] {e};

\node (etam0) [above=0.7cm of output_mix0]{};
\node (etam1) [above=0.6cm of output_mix1]{};
\node (mum0) [above=0.5cm of output_mean0]{};
\node (mum1) [above=0.4cm of output_mean1]{};
\node (sigmam0) [above=0.7cm of output_var0]{};
\node (sigmam1) [above=0.4cm of output_var1]{};
\node (eta0) [above=0.3cm of etam0]{$\hat{\eta}_{1, t+1}$};
\node (eta1) [above=0.4cm of etam1]{$\hat{\eta}_{2, t+1}$};
\node (mu0) [above=0.5cm of mum0]{$\hat{\mu}_{1, t+1}$};
\node (mu1) [above=0.6cm of mum1]{$\hat{\mu}_{2, t+1}$};
\node (sigma0) [above=0.3cm of sigmam0]{$\hat{\sigma}_{1, t+1}^2$};
\node (sigma1) [above=0.6cm of sigmam1]{$\hat{\sigma}_{2, t+1}^2$};

\draw[-] (input_bias.north) -- (hidden_mean0.south);
\draw[-] (input_bias.north) -- (hidden_mean1.south);
\draw[-] (input_bias.north) -- (hidden_mean2.south);
\draw[-] (input.north) -- (hidden_mean0.south);
\draw[-] (input.north) -- (hidden_mean1.south);
\draw[-] (input.north) -- (hidden_mean2.south);
\draw[-] (input_bias.north) -- (hidden_mix0.south);
\draw[-] (input_bias.north) -- (hidden_mix1.south);
\draw[-] (input_bias.north) -- (hidden_mix2.south);
\draw[-] (input.north) -- (hidden_mix0.south);
\draw[-] (input.north) -- (hidden_mix1.south);
\draw[-] (input.north) -- (hidden_mix2.south);
\draw[-] (var_input.north) -- (hidden_var0.south);
\draw[-] (var_input.north) -- (hidden_var1.south);
\draw[-] (var_input.north) -- (hidden_var2.south);
\draw[-] (var_input.north) -- (hidden_var3.south);
\draw[-] (var_input_bias.north) -- (hidden_var0.south);
\draw[-] (var_input_bias.north) -- (hidden_var1.south);
\draw[-] (var_input_bias.north) -- (hidden_var2.south);
\draw[-] (var_input_bias.north) -- (hidden_var3.south);
\draw[-] (hidden_input0.north) -- (hidden_var0.south);
\draw[-] (hidden_input0.north) -- (hidden_var1.south);
\draw[-] (hidden_input0.north) -- (hidden_var2.south);
\draw[-] (hidden_input0.north) -- (hidden_var3.south);
\draw[-] (hidden_input1.north) -- (hidden_var0.south);
\draw[-] (hidden_input1.north) -- (hidden_var1.south);
\draw[-] (hidden_input1.north) -- (hidden_var2.south);
\draw[-] (hidden_input1.north) -- (hidden_var3.south);
\draw[-] (hidden_mix_bias.north) -- (output_mix0.south);
\draw[-] (hidden_mix0.north) -- (output_mix0.south);
\draw[-] (hidden_mix1.north) -- (output_mix0.south);
\draw[-] (hidden_mix2.north) -- (output_mix0.south);
\draw[-] (hidden_mix_bias.north) -- (output_mix1.south);
\draw[-] (hidden_mix0.north) -- (output_mix1.south);
\draw[-] (hidden_mix1.north) -- (output_mix1.south);
\draw[-] (hidden_mix2.north) -- (output_mix1.south);

\draw[-] (hidden_mean_bias.north) -- (output_mean0.south);
\draw[-] (hidden_mean0.north) -- (output_mean0.south);
\draw[-] (hidden_mean1.north) -- (output_mean0.south);
\draw[-] (hidden_mean2.north) -- (output_mean0.south);
\draw[-] (hidden_mean_bias.north) -- (output_mean1.south);
\draw[-] (hidden_mean0.north) -- (output_mean1.south);
\draw[-] (hidden_mean1.north) -- (output_mean1.south);
\draw[-] (hidden_mean2.north) -- (output_mean1.south);

\draw[-] (hidden_var0.north) -- (output_var0.south);
\draw[-] (hidden_var0.north) -- (output_var1.south);
\draw[-] (hidden_var1.north) -- (output_var0.south);
\draw[-] (hidden_var1.north) -- (output_var1.south);
\draw[-] (hidden_var2.north) -- (output_var0.south);
\draw[-] (hidden_var2.north) -- (output_var1.south);
\draw[-] (hidden_var3.north) -- (output_var0.south);
\draw[-] (hidden_var3.north) -- (output_var1.south);
\draw[->] (output_mix0.north) -- (eta0.south);
\draw[->] (output_mix1.north) -- (eta1.south);
\draw[->] (output_mean0.north) -- (mu0.south);
\draw[->] (output_mean1.north) -- (mu1.south);
\draw[->] (output_var0.north) -- (sigma0.south);
\draw[->] (output_var1.north) -- (sigma1.south);
\draw[->] [sharp corners](output_mix0.north) -- (etam0.south) -- (8.82, 4.25) -- (sum.north);
\draw[->] [sharp corners](output_mix1.north) -- (etam1.south) -- (8.82,4.15) -- (sum.north);
\draw[->] [sharp corners](output_mean0.north) -- (mum0.south) -- (8.82,4.05) -- (sum.north);
\draw[->] [sharp corners](output_mean1.north) -- (mum1.south) -- (8.82,3.95) -- (sum.north);
\draw[->] [sharp corners](sum.south) -- (var_input);
\draw[->] [sharp corners](input.east) -- (var_input);
\draw[->] [sharp corners](output_var0.north) -- (sigmam0.south) -- (16.6,4.25) -- (16.6, -1.3) -- (13.45, -1.3) -- (hidden_input0.south);
\draw[->] [sharp corners](output_var1.north) -- (sigmam1.south) -- (16.3,3.95) -- (16.3, -1) -- (14.95, -1) -- (hidden_input1.south);
\matrix [column sep=0.25cm, below, ampersand replacement=\&,column1/.style={anchor=current east},
    column2/.style={anchor=current east}] at (current bounding box.south) {
  \node{s: softmax}; \&  \node{$\Sigma$: linear activation function}; \\
 `\node{g: hyperbolic tangent activation function}; \& \node{e: positive exponential linear unit}; \\
};

\end{tikzpicture}}
    \caption{Architecture of the proposed RMDN}
    \label{fig:architecture}
\end{figure}
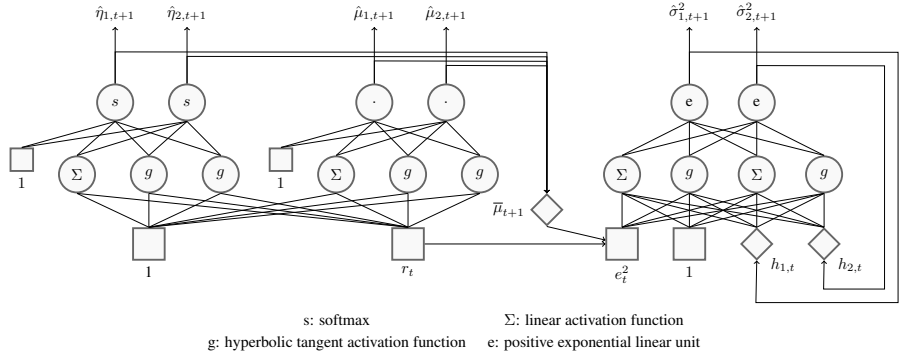

The architecture of the ELU-RMDN can be separated into three parts, which are used to estimate the parameters of a mixture of $N$ components. For simplicity, we restrict the model to a lag of 1 for neural network inputs. In this paper, we only consider a mixture of Gaussian distributions. The mixture distribution is characterized by three sets of parameters: the weight of each of the components $\hat{\eta}_{t+1} = (\hat{\eta}_{1,t+1},\hat{\eta}_{2,t+1}, ..., \hat{\eta}_{N,t+1})$, their conditional mean $\hat{\mu}_{t+1} = (\hat{\mu}_{1,t+1}, \hat{\mu}_{2,t+1}, ..., \hat{\mu}_{N,t+1})$ and their conditional variance $\hat{\sigma}_{t+1}^2 = (\hat{\sigma}_{1,t+1}^2, \hat{\sigma}_{2,t+1}^2, ..., \hat{\sigma}_{N,t+1}^2)$. The conditional mean and variance of the mixture distribution are respectively given by $\overline{\mu}_{t+1} = \sum_{i=1}^N \hat{\eta}_{i,t+1}\hat{\mu}_{i,t+1} $ and $\overline{\sigma}_{t+1}^2 = \sum_{i=1}^N \hat{\eta}_{i,t+1}(\sigma_{i,t+1}^2 + (\hat{\mu}_{i,t+1}-\overline{\mu}_{t+1})^2)$. The input of the ELU-RMDN is a time series $\{r_t\}_{t=1}^T$. The first part of the ELU-RMDN is the mixing network, which takes for input $r_{t}$  and estimates the weights $\hat{\eta}_{t+1}$ of the mixture components. 
The output layer of this network is followed by a softmax activation function $s_i(y_1,\ldots,y_N) =  \frac{e^{y_i}}{\sum_{j=1}^N e^{y_j}}$ to ensure that the condition $\sum_{i=1}^N \hat{\eta}_{i, t+1}=1$ is satisfied. The estimate $\hat{\eta}_{t+1}$ is thus defined as
\begin{equation}
    \hat{\eta}_{i, t+1}= s_i \left(u_{n, 1} (U_{1,1}r_t +U_{1,0})+\sum_{k=2}^K u_{n, k} tanh(U_{k, 1} r_t +U_{k,0}) + u_{n, 0}, \, n=1,\ldots,N \right)
\end{equation}
where $U$ are the input-to-hidden layer weights and $u$ are the hidden-to-output layer weights. The second part of the model is the mean-level network, which takes as input $r_{t}$ and estimates the mean $\hat{\mu}_{i,t+1}$ for each mixture component $i$. The estimate $\hat{\mu}_{t+1}$ is thus defined through 
\begin{equation}
    \hat{\mu}_{i, t+1}= v_{i, 1} (V_{1,1}r_t +V_{1,0})+ \sum_{k=2}^K v_{i, k} tanh(V_{k, 1} r_t +V_{k,0})+ v_{i, 0}
\end{equation}
where $V$ are the input-to-hidden layer weights and $v$ are the hidden-to-output layer weights. The third part of the ELU-RMDN is the variance recurrent network, which estimates $\hat{\sigma}_{t+1}^2$, the conditional variance of the various components of the mixture. As in the GARCH model, this network takes for input the residual $e_{t}^2 = (r_t - \overline{\mu}_{t})^2$ and the last estimate for the conditional variance $\hat{\sigma}_{t}^2$. The output layer is fed through what we refer to as the positive exponential linear unit, which is defined as $e(x, \alpha) = ELU(X, \alpha) +1 +\epsilon$ where epsilon is a very small number and $ELU(x, \alpha)$ is the exponential linear unit defined as 
\begin{equation}
    ELU(x, \alpha) = \begin{cases}
        x & \text{if } x >0\\
        \alpha (e^x - 1) & \text{otherwise}
    \end{cases}
\end{equation} 
The hyperparameter $\alpha$ controls the saturation value for negative inputs\citep{elu}. The ELU activation function improves the numerical stability of MDN and ensures that the estimated variance is always greater than 0\citep{Guillaumes2017MixtureDN}\citep{deepandshallow}. We refer to this activation function as the "positive exponential linear activation unit" throughout this paper. The use of such activation function is what differentiates our architecture from that of \cite{Nikolaev2012}. The estimate for $\hat{\sigma}_{i,t+1}$ is 
\begin{align}
    \hat{\sigma}_{i,t+1}&= e \bigg(w_{i, 1} (W_{1,1}e_{t}^2 +W_{1,0})+ \sum_{k=2}^K w_{i, k} tanh(W_{k, 1} e_t^2 +W_{k,0})\\
    \nonumber &+ w_{i, K+1} (W_{K+1,1}\hat{\sigma}_{i,t}^2 +W_{K+1,0})+ \sum_{k=K+2}^{2K} w_{i, k} tanh(W_{k, 1} \hat{\sigma}_{i,t}^2  +W_{k,0})+ w_{i, 0} \bigg)
\end{align}
where $w$ are the hidden-to-output weights and $W$ are the input-to-hidden weights. 
\footnote{Note that to address identifiability issues, some of the weights can be set to either $0$ or $1$ without loss of generality. For instance, we can set $U_{1,1}=V_{1,1}=W_{1,1}=1$ and $U_{1,0}=V_{1,0}=W_{1,0}=0$.}
A detailed graphical representation of the architecture is presented in Figure \ref{fig:architecture}. 
It is worth noting that using a linear node in the hidden layer implies that the AR(1)-GARCH(1, 1) model is a particular case of the ELU-RMDN model. Consider a AR(1)-GARCH(1, 1) defined as 
\begin{align}
r_{t+1} & = \mu_{t+1} + \sigma_{t+1} e_{t+1} \\
\nonumber\mu_{t+1} &= a_0 + a_1 r_t\\
\nonumber\sigma_{t+1}^2 &= \alpha_0 + \alpha_1 e_{t}^2 +\beta_1 \sigma_{t}^2
\end{align}
where $e_t\sim N(0,1)$ denotes an innovation. Now consider an ELU-RMDN with only $N=1$ component and with only the linear nodes in the hidden layer of each network. We achieve this by setting $V_i=0$ for $1<i \leq K$ and $W_i = 0$ for $1<i \leq K$ and for $K+2<i \leq 2K$.
Since only one component is considered, we can ignore the mixture weights $\hat{\eta}_{t+1}$. The RMDN-GARCH can then be reformulated mathematically as 
\begin{align}
    r_{t+1} &= \mu_{t+1} +\sigma_{t+1}e_{t+1}\\
    \nonumber\hat{\mu}_{t+1} &= v_1 (V_{1}r_t +V_{0}) +v_0\\
    \nonumber\hat{\sigma}_{t+1}^2&= e(w_{1} (W_{1}e_{t}^2 +W_{0}) + w_{2} (W_{2}\hat{\sigma}_{t}^2 +W_{3}) +w_{0})
\end{align}
Such representation of the variance $\hat{\sigma}_{t+1}^2$ is equivalent to conditional variance of a GARCH(1,1) when the input of the ELU is greater than $0$. The conditional mean also $\hat{\mu}_{t+1}$ simplifies to an expression equivalent to an AR(1) model. This implies that the proposed ELU-RMDN reduces to an AR(1)-GARCH(1,1) when the input of the ELU is greater than 0.

In this paper the model is implemented using PyTorch \citep{paszke2017automatic}. The ELU-RMDN is trained by minimizing the negative log-likelihood. Assuming a mixture of $N$ Gaussian components and denoting by $\phi(\cdot,\mu,\sigma^2)$ the density of a Gaussian distribution with mean $\mu$ and variance $\sigma^2$, the negative log-likelihood $-\ell$ is
\begin{eqnarray}
-\ell  &=& \sum_{t=1}^T -log\;\sum_{i=1}^N \hat{\eta}_{i,t}\phi(r_t,\hat{\mu}_{i,t}, \hat{\sigma}_{i,t}^2) \notag
\\ &=& \sum_{t=1}^T -log\;\sum_{i=1}^N exp \biggr[ log\; \hat{\eta}_{i,t}- \frac{1}{2} log\;2\pi - log\;\hat{\sigma}_{i, t} -  \frac{1}{2} \frac{(r_t - \hat{\mu}_{i,t})^2}{\hat{\sigma}_{i, t}^2}\biggr]. \label{loglikreformul}
\end{eqnarray}
The formulation \eqref{loglikreformul} makes it possible to use the \textit{logsumexp} trick from \cite{Blanchard2016} to avoid underflow issues.
We use the \textit{logsumexp} implementation available in the PyTorch library \citep{paszke2017automatic} as a direct replacement of the expression $log \sum_{i=1}^N exp $ in the loss function.
We train the model using the Adam optimizer of \citep{Adam2014} without weight regularization.

\section{Pretraining Method}
As previously noted, a well-known problem with mixture density networks is their propensity to hit (potentially bad) local minima\citep{hepp_mixture_2022}. This section details a method for pretraining the ELU-RMDN, which makes it possible to avoid local minima that would result in a log-likelihood lower than that of the GARCH model. Indeed, as mentioned above, the ELU-RMDN model is a generalization of the Gaussian AR-GARCH model, with the latter being nested within the former. A consequence of this relationship between the GARCH and the ELU-RMDN is that the optimized log-likelihood of the ELU-RMDN should always be greater or equal to that of a AR-GARCH.\footnote{Although the ELU-RMDN of this paper is designed to nest the AR(1)-GARCH(1), the method proposed can easily generalize to encompass AR(p)-GARCH(P,Q) models.} However, in early experiment the ELU-RMDN did not always converge to a minimum with a log-likelihood greater to that of the GARCH. The model was particularly unstable and exhibited the so-called \textit{persistent NaN problem} often encountered with mixture density networks, as defined by \cite{Guillaumes2017MixtureDN}. 

We propose a pretraining for the ELU-RMDN that makes use of the linear nodes of the hidden layer of each neural network. The intuition behind our approach is that the parameters of a linear ELU-RMDN should be a good starting point for the parameters of a non-linear ELU-RMDN. The pretraining of the model starts by freezing the non-linear nodes of the hidden layer of each neural network. This is accomplished by setting the gradients of the node with non-linear activation function on the hidden layer to 0 every time the gradient is computed. The model is then trained using all the nodes in the hidden layer.
\section{Performance Evaluation}
We present numerical experiments comparing the performance of the proposed training approach based on the proposed linear pretraining to that of an ELU-RMDN that is initialized randomly and that is not pretrained. 
\subsection{Methodology}
The performance of the two training methods is evaluated on daily returns of 10 stocks selected randomly from the S\&P 500 universe. We train the model on the period extending between September 20, 2017 and September 10, 2021. We first initialize the all of the ELU-RMDN parameters randomly, with the exception of parameters of the output node of the variance recurrent network and the biases of all networks, which were initialized to 1. However, as the pretraining method only trains the linear node of the neural network, we initialize the weights of the non-linear nodes to zeros. This step ensures that the model converges to the optimal linear model.

Both models are initialized randomly using 10 different seeds (one per training run) selected randomly between 0 and 50000. For the pretrained ELU-RMDN, we pretrain the ELU-RMDN for 20 epochs and then train all the nodes for 300 epochs. The ELU-RMDN without pretraining is trained for 300 epochs. The performance of the two training methods is evaluated by splitting the training runs for each stock into 2 groups, which we define as "Not Converged" and "Converged". The "Not Converged" group consists of the group of training runs for which the log-likelihood is equal to NaN or is unreasonable (smaller than -$100,\!000$). NaN values are due to the ELU-RMDN evaluating the parameters of the mixture to NaN, which can occur for instance when the gradient explodes. The "Converged" group is comprised of the remainder of the training runs, which exhibit a reasonable log-likelihood. We also evaluate the average log-likelihood across runs after training for each stock for a GARCH(1,1) with an underlying Gaussian distribution and for the ELU-RMDN with and without the pretraining.
\subsection{Results}
We report the convergence status for all the training groups in Table \ref{tab:is_table}. We report the average log-likelihood in Table \ref{tab:in_sample_nll}.
\begin{table}[!ht]
\begin{adjustbox}{width=0.6\textwidth, center}
\begin{tabular}[c]{ccccc}
\cline{2-5}
  \multicolumn{1}{c}{ }& \multicolumn{2}{c}{Without pretraining} & \multicolumn{2}{c}{With pretraining} \\
  \multicolumn{1}{c}{Stock Ticker }& \multicolumn{1}{p{1.4cm}}{\centering Not Converged} & \multicolumn{1}{p{1.4cm}}{\centering  Converged} & \multicolumn{1}{p{1.4cm}}{\centering Not Converged} & \multicolumn{1}{p{1.4cm}}{\centering Converged}\\
 \hline
AKAM & 7 & 3 & 0 & 10\\
CBRE & 6 & 4 & 0 & 10\\
EA & 7 & 3 & 0 & 10\\
EMN & 6 & 4 & 0 & 10\\
K & 7 & 3 & 0 & 10\\
LOW & 7 & 3 & 0 & 10\\
MKTX & 7 & 3 &  0 & 10\\
NWSA & 7 & 3 &  0 & 10\\
OKE & 7 & 3 & 0 & 10\\
UAL & 8 & 2 & 0 & 10\\
\hline
Total & 69 & 31 & 0 & 100\\
Total\% & 69\% & 31\% & 0\% & 100\%\\
\hline
\end{tabular}
\end{adjustbox}
\caption{Results of In-Sample Convergence Tests across 10 stocks}
\label{tab:is_table}
\end{table}
The results show that the ELU-RMDN with pretraining converges in all cases, while the ELU-RMDN without pretraining does not converge in the majority of cases. The results presented in Table \ref{tab:in_sample_nll} show that using the pretraining leads in average to a likelihood greater than that of the GARCH model, with the exception of the training runs performed on the EMN stock. However, after retraining the model on the EMN return series with a smaller learning rate, the model was able to converge well below the log-likelihood of the GARCH. This suggests that the choice of learning rate is also an important factor related to the convergence of the ELU-RMDN.
\begin{table}[!ht]
\centering
    \begin{tabular}{lrrr}
          \cline{2-4}
          \multicolumn{1}{b{5.5em}}{\parbox[t]{5.5em}{\centering Stock\\Ticker}}& \multicolumn{1}{b{5.5em}}{\parbox[t]{5.5em}{\centering \quad\\GARCH}} & \multicolumn{1}{b{5.5em}}{\parbox[t]{5.5em}{\centering With \\Pretraining}} & \multicolumn{1}{b{5.5em}}{\parbox[t]{5.5em}{\centering Without \\Pretraining}} \\
          \hline
    AKAM  & -1999.45 & -1875.84 & -1898.40 \\
    CBRE  & -1973.91 & -1934.19 & -1951.72 \\
    EA    & -2038.76 & -1995.17 & -2078.30 \\
    EMN   & -2025.97 & -2027.82 & -2033.02 \\
    K     & -1782.75 & -1695.07 & -1700.97 \\
    LOW   & -1993.17 & -1906.67 & -1914.93 \\
    MKTX  & -2071.61 & -2022.83 & -2040.89 \\
    NWSA  & -2041.73 & -1968.67 & -1989.55 \\
    OKE   & -2054.26 & -2006.29 & -2022.70 \\
    UAL   & -2349.71 & -2303.32 & -2335.57 \\
    \hline
    \end{tabular}%

\caption{Average Log-Likelihood across 10 stocks}
\label{tab:in_sample_nll}
\end{table}

\section{Conclusion and Future Work}
In this paper, we propose a novel method for training a recurrent mixture density network. Through empirical analysis on stock return data, we show that our pretraining method improves the robustness of our model to bad local minima. Our results also show that the ELU-RMDN trained using our pretraining method does not suffer from frequently obtaining NaN values, which are often obtained in absence of pretraining. Further research could focus on the application of pretraining methods to other neural networks for which a linear counterpart model exists, such as the autoencoder whose linear counterpart is principal component analysis (PCA).
\bibliographystyle{unsrtnat}
\bibliography{references}

\end{document}